\documentclass{article}

\usepackage{arxiv}
\usepackage{natbib}
\setcitestyle{numbers,square}
\usepackage[utf8]{inputenc} 
\usepackage[T1]{fontenc}    
\usepackage{hyperref}       
\usepackage{url}            
\usepackage{booktabs}       
\usepackage{amsfonts}       
\usepackage{nicefrac}       
\usepackage{microtype}      
\usepackage{lipsum}
\usepackage{graphicx}
\usepackage{amsmath}
\usepackage{indentfirst}
\graphicspath{ {./images/} }

\title{TKFNet: Learning Texture Key Factor Driven Feature for Facial Expression Recognition}

\author{
 Liqian Deng \\
  National Engineering Research Center of E-Learning  \\
  Central China Normal University\\
  Wuhan, China \\
  \texttt{denglichien@mails.ccnu.edu.cn} \\}
\date{}

\begin{document}
\maketitle

\begin{abstract}
Facial expression recognition (FER) in the wild remains a challenging task due to the subtle and localized nature of expression-related features, as well as the complex variations in facial appearance. In this paper, we introduce a novel framework that explicitly focuses on Texture Key Driver Factors (TKDF), localized texture regions that exhibit strong discriminative power across emotional categories. By carefully observing facial image patterns, we identify that certain texture cues, such as micro-changes in skin around the brows, eyes, and mouth, serve as primary indicators of emotional dynamics. To effectively capture and leverage these cues, we propose a FER architecture comprising a Texture-Aware Feature Extractor (TAFE) and Dual Contextual Information Filtering (DCIF). TAFE employs a ResNet-based backbone enhanced with multi-branch attention to extract fine-grained texture representations, while DCIF refines these features by filtering context through adaptive pooling and attention mechanisms. Experimental results on RAF-DB and KDEF datasets demonstrate that our method achieves state-of-the-art performance, verifying the effectiveness and robustness of incorporating TKDFs into FER pipelines.
\end{abstract}


\section{Introduction}

Facial Expression Recognition (FER) is an essential branch of emotion understanding. FER focuses on detecting and interpreting human emotions through facial movements. This technique has wide-ranging applications across multiple domains, including education\citep{guoFacialExpressionsRecognition2022}, human-computer interaction\citep{chowdaryDeepLearningbasedFacial2023}, mental health assessment\citep{j.yeDepFERFacialExpression2024}. Although FER systems have shown promising results under controlled conditions, deploying them in real-world environments remains challenging. Variability in lighting, head orientation and partial occlusions can all compromise recognition accuracy. As depicted in Fig. 1, facial features may be distorted or concealed due to poor lighting, side profiles, or blocked regions. These challenges underscore the need for more robust and adaptable FER systems.

\begin{figure}[h] 
	\centering 
	\includegraphics[scale=0.4]{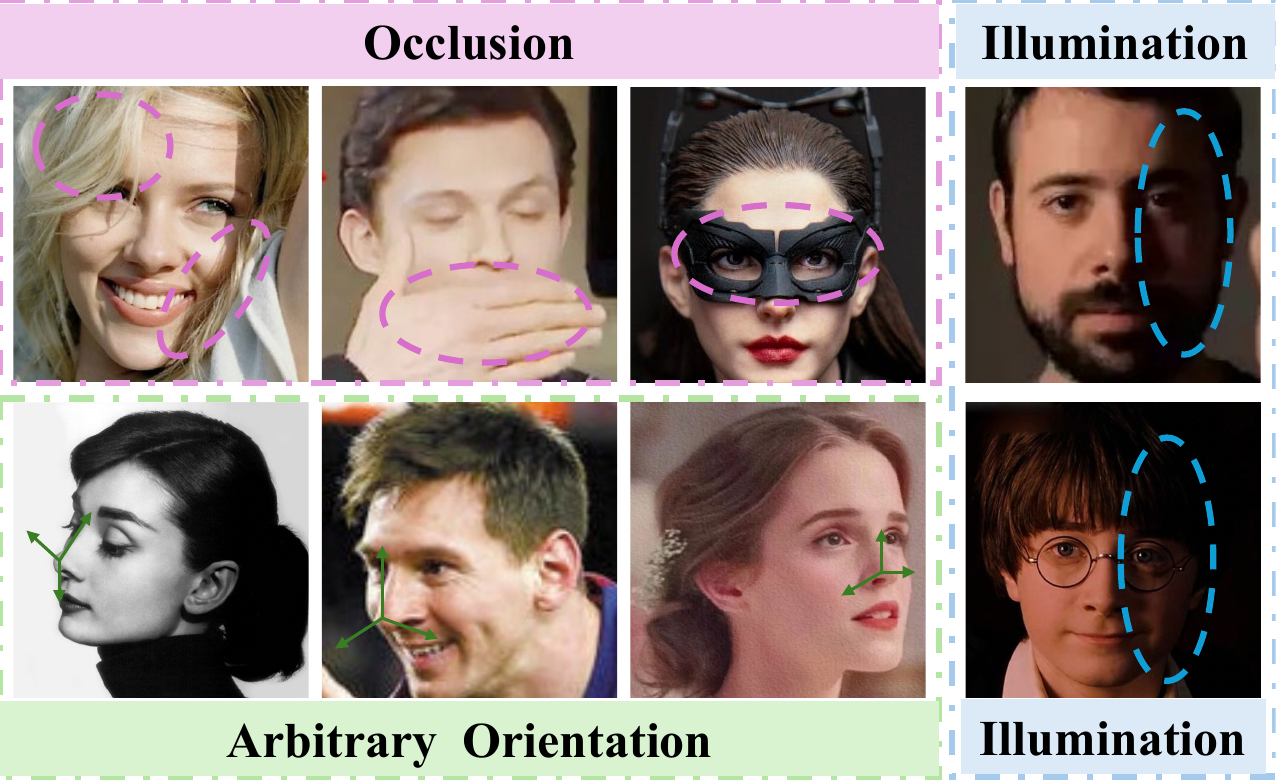}
	\caption{Challenges in FER, including arbitrary orientations, illumination and occlusion.} 
	\label{FIG:1} 
\end{figure}

\par{By carefully observing facial images, we identify Texture Key Driver Factors(TKDF) that play a crucial role in the dynamics of facial expression changes. It refers to a local texture region or texture descriptor that significantly captures differences among emotional categories and exhibits high discriminative power in facial expression recognition. These factors serve as the core driving elements behind subtle variations in facial expressions and provide critical cues for distinguishing between different emotions. As illustrated in Fig. 2, in the happy expression shown in Fig. 2(a), the key driver factors for the eyes include the texture changes leading to narrowed eyes, while the key driver factors for the mouth involve features such as smile lines. Similarly, in Fig. 2(b), eyebrow factors contribute to the frown expression. In the case of surprise, both the eyes and mouth are influenced by specific factors. These key driver factors guide the model in focusing on the most informative features within facial images.}

\begin{figure}[h] 
	\centering 
	\includegraphics[scale=0.35]{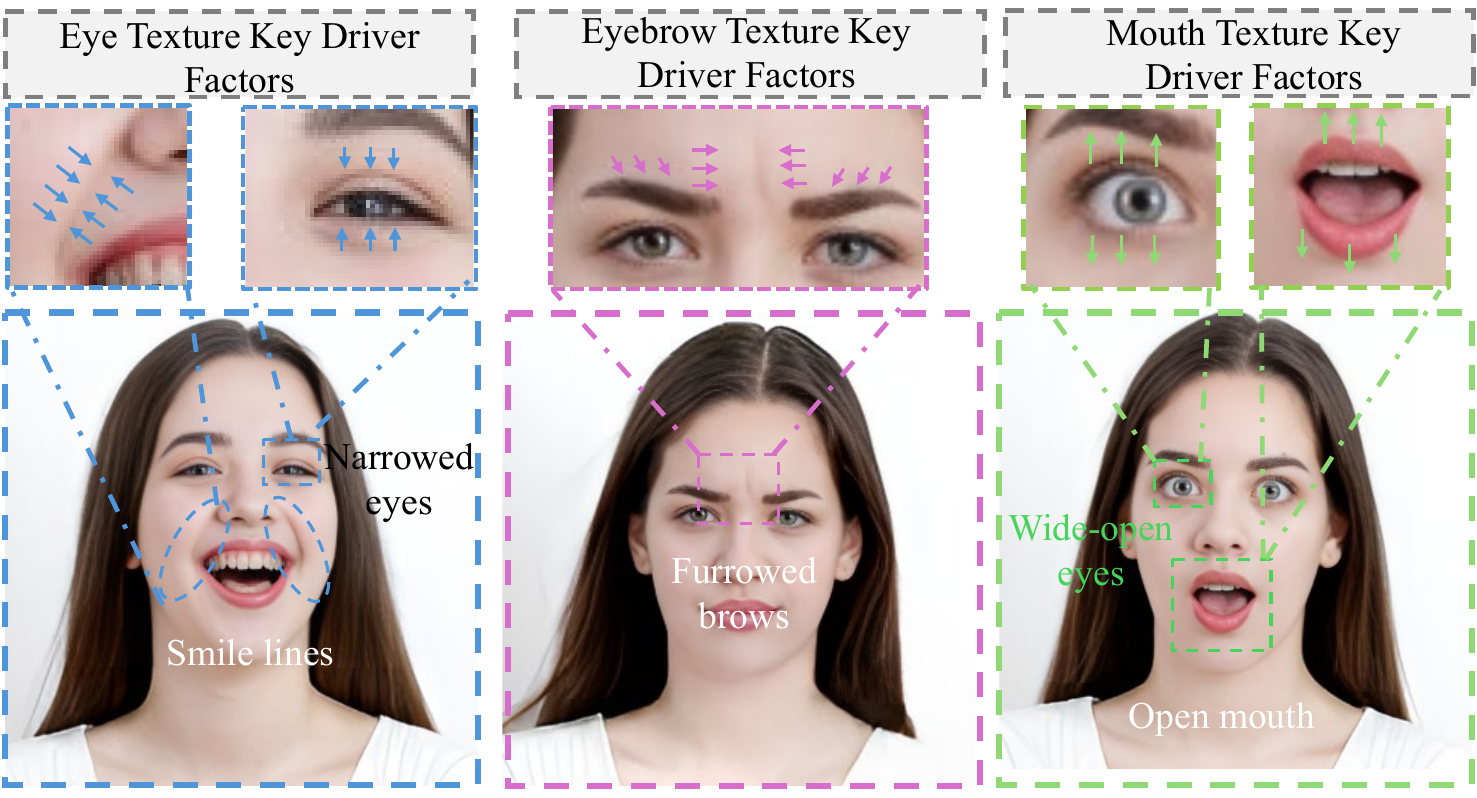}
	\caption{Texture Key driven factors} 
	\label{FIG:2} 
\end{figure}

\par{In this work, we propose a novel method that effectively mines texture key driving factors and leverages them to enhance the discovery of discriminative features for subtle facial expression recognition. Specifically, our model is composed of two key components: the Texture-Aware Feature Extractor (TAFE) and the Dual Contextual Information Filtering (DCIF) module. The proposed TAFE and DCIF modules work collaboratively, where TAFE focuses on extracting fine-grained texture cues via ResNet and statistical attention modeling, while DCIF selectively filters contextual information through adaptive pooling and attention mechanisms, together enhancing the model’s sensitivity to subtle variations and its robustness across complex expression distributions. In summary, the major contribution of this work are listed below:}

\begin{itemize}
\item We identify and leverage texture key driving factors that play a pivotal role in facial expression recognition.
\item We propose a two-branch architecture combining TAFE and DCIF to effectively extract and refine discriminative features through multi-scale texture modeling and contextual filtering.
\item Extensive experiments on RAF-DB and KDEF demonstrate the performance of our method compared to state-of-the-art approaches, validating its robustness and generalization capability.
\end{itemize}

\section{Method}
\subsection{Overview}
\par{Figure 3 illustrates the overall architecture, which integrates two core components: TAFE and DCIF. TAFE employs a ResNet backbone to capture fine-grained local skin texture cues that are crucial for subtle facial expression recognition, enhancing the model’s sensitivity to low-level semantic variations. DCIF incorporates attention mechanisms to effectively filter and emphasize the most informative features, enabling the model to focus on contextually relevant cues for more accurate expression recognition.}

\begin{figure}[h] 
	\centering 
	\includegraphics[scale=0.55]{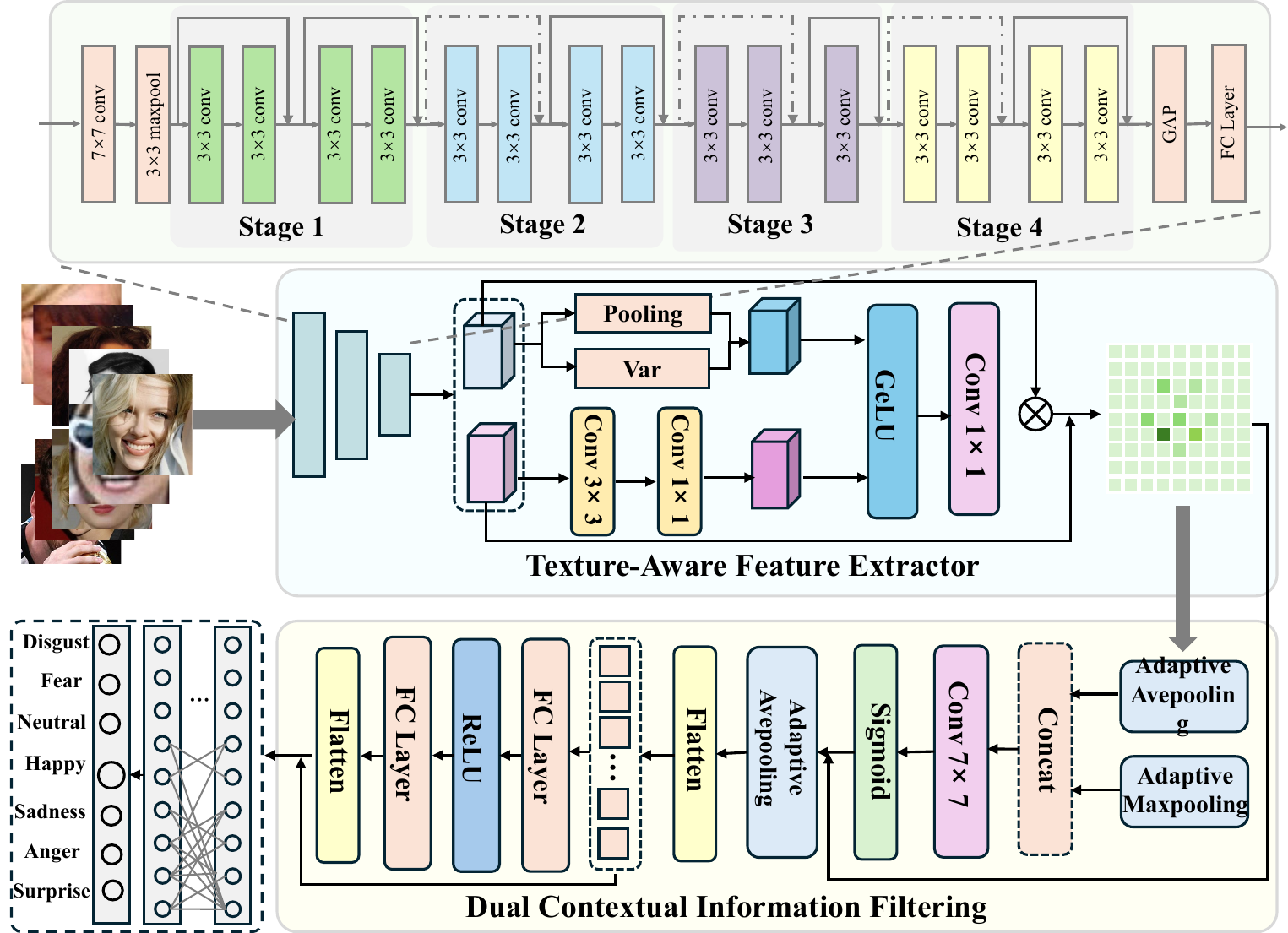}
	\caption{Pipeline of TKFNet.} 
	\label{FIG:3} 
\end{figure}

\subsection{Texture-Aware Feature Extractor}
\par{Given a batch of facial expression images $F=\{(e_{1},l_{1}),(e_{2},l_{2}),...,(e_{b},l_{b})\}$, where $e_{i}\in R_{H\times W\times C}$ denotes an input image and $l_{i}$ represents the corresponding label. We first extract deep features from each image using a ResNet backbone which is effective at capture local features:
\begin{equation}
   \varphi_i=g\left(e_i\right),
\end{equation}
where $g(*)$ denotes the feature extraction function implemented by the ResNet model. To capture distinct texture-sensitive representations, we introduce a dual-branch structure to disentangle different types of local texture information. This is achieved by passing $\varphi_{i}$ into two separate branches as follow:}
\begin{equation}
\begin{cases}
	\mathbf{o}_1=\mathbf{W}_1\cdot\varphi_i(h,w,:)+\mathbf{b}_1 \\
	\mathbf{o}_2=\mathbf{W}_2\cdot\varphi_i(h,w,:)+\mathbf{b}_2 & 
\end{cases}\text{for all }(h,w).
\end{equation}
\par{For the first branch $o_{1}$,  we aim to generate a fine-grained attention modulation that is sensitive to local texture variations. To this end, we design a texture-aware descriptor by integrating both semantic and statistical cues from the feature map. Specifically, we compute two complementary representations:
\begin{equation}
\begin{cases}
	\mathbf{O}_s=\frac{1}{HW}\sum_{h=1}^{H}\sum_{w=1}^{W}o_1(h,w,:) \\
	\mathbf{O}_v=\frac{1}{HW}\sum_{h=1}^{H}\sum_{w=1}^{W}\left(o_1(h,w,:)-\mathbf{O}_s\right)^2 & 
\end{cases}.
\end{equation}}
\par{We then linearly combine these two descriptors to form a unified modulation signal:
\begin{equation}
O=\alpha O_s+\beta O_v,
\end{equation}
where $alpha$ and $beta$ are learnable scalar weights that adaptively balance the contribution of semantic and statistical information. This fused representation $O$ is subsequently used to recalibrate the original feature map via channel-wise multiplication, allowing the model to emphasize informative texture channels while suppressing irrelevant or redundant ones.}
\par{Then, we refine the feature map by applying a non-linear activation followed by convolution, and reweight it with the original feature via element-wise multiplication:
\begin{equation}
    \vartheta^1=Conv\left(\sigma(O)\right)\odot o_1,
\end{equation}
where $\sigma(*)$ represents GeLU activation function. The symbol $\odot$ represents the Hadamard product (i.e., element-wise multiplication), which performs channel-wise modulation of the feature map by applying weights.}
\par{For the second branch $o_{2}$, we aim to enhance its capacity to capture contextual texture patterns by employing a cascaded convolutional structure. This design enables the model to integrate local receptive fields with non-linear transformations, thereby capturing rich spatial dependencies and subtle texture details. Specifically, we apply a sequence of convolutions as follows:}
\begin{equation}
    \vartheta^2=Conv_{1\times1}\left(\sigma\left(Conv_{1\times1}(Conv_{3\times3}(o_2))\right)\right).
\end{equation}
\par{Finally, we concatenate the two refined branches to obtain the fused texture-aware representation:}
\begin{equation}
    \vartheta^{\prime}=Concat(\vartheta^1,\vartheta^2).
\end{equation}

\subsection{Dual Contextual Information Filtering}
\par{After obtaining the enhanced feature map $\vartheta^{\prime}$, we proceed to adaptive pooling through two parallel branches to capture global context information from different aspects of the feature map. This step is crucial for the following reason:}

\begin{equation}
\begin{cases}
	r_1=AdpAvePooling(\vartheta^{\prime}) \\
	r_2=AdpMaxPooling(\vartheta^{\prime}) & 
\end{cases}.
\end{equation}
    
\par{By concatenating these two pooled representations $r_1$ and $r_2$, we effectively combine both global and local information into a single rich descriptor $R$. }
\begin{equation}
   R=Concat(r_1,r_2)
\end{equation}

\par{Next, we process the concatenated vector R through a lightweight convolutional neural network (CNN) followed by a sigmoid activation function to learn the optimal feature combination and to generate an attention map $A$ that reflects the relative importance of each feature. Then we apply it on feature map $vartheta^{\prime}$ via element-wise multiplication.}
\begin{equation}
\eta=Sigmoid\left(Conv(R)\right),
\end{equation}
\begin{equation}
    \theta=\eta\odot\vartheta^{\prime}.
\end{equation}

\par{To further enhance the representation power of the feature map $\theta$, we apply a compact and efficient global context encoding mechanism. This process aims to capture holistic information that reflects the overall distribution of expression-relevant features across the entire image. First, we apply adaptive average pooling, which reduces the spatial dimensions while preserving the global contextual patterns. This operation condenses the feature map into a compact summary vector:
\begin{equation}
    \kappa_1=F\left(AdpAvePooling(\theta)\right),
\end{equation}
where $F(*)$ represents the flatten operation.}
\par{Here, $kappa_{1}$ is a flattened feature vector representing the global average statistics of the input feature map. It serves as a lightweight yet informative global descriptor. Next, we pass $kappa_{1}$  through a two-layer fully connected network with a ReLU activation in between. This non-linear transformation enables the network to project the pooled global features into a more expressive latent space:

\begin{equation}
    K=F\left(FC\left(\rho (FC(\kappa_1))\right)\right),
\end{equation}
where $\rho(*)$ denotes the ReLU activation function.}

\par{The resulting vector $K$ captures a richer semantic representation of the global facial context, which can be further used for tasks such as attention generation, expression classification, or feature refinement. This mechanism helps the model become more sensitive to subtle differences in facial structure and emotional cues that are not always localized.}

\par{Finally, we gain the final output as follows:}
\begin{equation}
    Logits=FC\left(Flatten\left(Gap(K)\right)\right)
\end{equation}

\subsection{The total loss function}
In our approach, we adopt the Cross-Entropy Loss function, which is widely recognized as an effective objective function for multi-class classification tasks. It measures the divergence between the predicted probability distribution output by the model and the actual ground truth labels, guiding the model to make more accurate predictions through iterative optimization. The Cross-Entropy Loss can be formally expressed as:
\begin{equation}
L_{total}=-\frac{1}{N}\sum_{k}\sum_{c=1}^{Q}y_{kc}log(p_{kc}),
\end{equation}
where $Q$ denotes the total number of classes, $y_{kc}$ is the symbolic function (with a value of 1 if the sample belongs to the $c$-th class and 0 otherwise), and $p_{kc}$ refers to the model’s predicted probability that sample $k$ belongs to class $c$. This loss function penalizes incorrect predictions by taking the negative logarithm of the predicted probability assigned to the true class, thereby encouraging the model to assign higher confidence scores to correct classifications. The total loss is averaged over all samples in the dataset to ensure stable gradient updates during training.

\section{Experiments}
\par{In this section, we provide a detailed introduction to the two datasets used in our experiments, followed by the experimental setup and a comprehensive presentation of the results. The evaluation includes comparisons with state-of-the-art methods, along with visualizations that demonstrate the effectiveness and predictive performance of our model.}
\subsection{Dataset}
\par{The RAF-DB (Real-World Affective Faces Database)\citep{s.liReliableCrowdsourcingDeep2017a} is a comprehensive dataset containing over 30,000 facial images, each annotated with one of seven basic emotions. Emphasizing spontaneous expressions captured in real-life situations, it serves as a valuable resource for emotion recognition across diverse environments. In contrast, the KDEF (Karolinska Directed Emotional Faces) \citep{calvoFacialExpressionsEmotion2008}dataset features high-quality images of 70 individuals, each portraying seven distinct emotions under controlled conditions. Its consistency and clarity make it a popular choice in facial expression and psychological research. The samples is shown in Fig. 4.} 
\begin{figure}[h] 
	\centering 
	\includegraphics[scale=0.48]{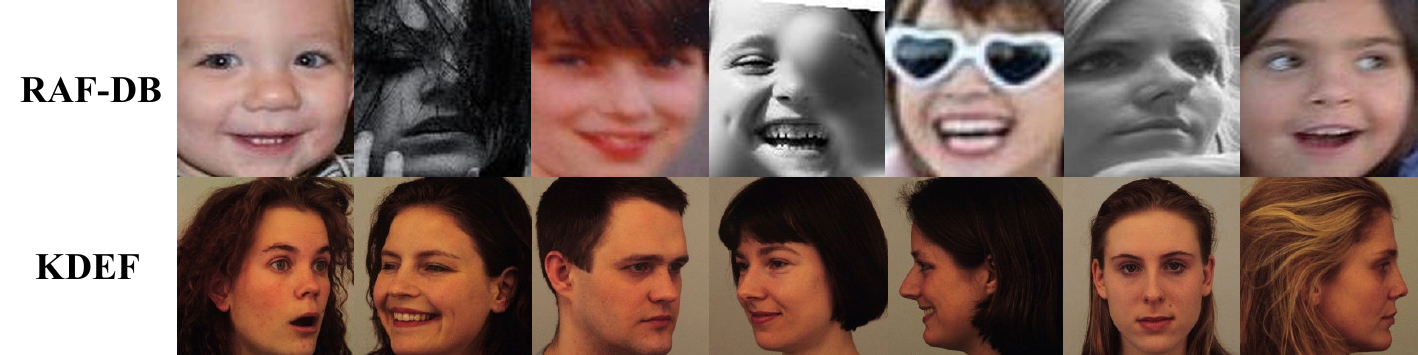}
	\caption{Samples in RAF-DB and KDEF datastes.} 
	\label{FIG:4} 
\end{figure}
\subsection{Experiment details}
\par{In our experimental pipeline, facial images are first automatically detected and cropped to isolate expression-relevant regions. These cropped images are then uniformly resized to 224 $\times$ 224 pixels to satisfy the input size requirements of the neural network. The model is trained for 60 epochs with a batch size of 128. We employ a Momentum optimizer with an initial learning rate of 0.1, enhanced by a polynomial decay strategy that gradually reduces the learning rate to 0.01 over a predefined number of steps, using a decay power of 0.5.  All model development and training procedures are implemented using the MindSpore framework. Experiments are conducted on a computing platform equipped with an NVIDIA T4 GPU.}
\subsection{Comparison with state-of-the-art methods}	
\par{We evaluate our method by comparing it with state-of-the-art approaches on both the RAF-DB and KDEF datasets. The results show that our method achieves superior performance, surpassing the latest techniques in the field.}

\par{\textbf{(1) Results on RAF-DB.} As shown in Table 1, the TKFNet we proposed achieved a recognition accuracy rate of 85.XX \% on the RAF-DB dataset, demonstrating a relatively good performance advantage. Among them, HealthFERS is still slightly lower than our model, verifying the effectiveness of the texture-aware dual-branch architecture we proposed in the expression recognition task. In contrast, EQCNN achieved accuracy rates of 81.95\% . Although they also have certain competitiveness, they have deficiencies when dealing with complex facial texture changes. Meanwhile, the traditional attention mechanism method only achieved an accuracy rate of 81.09\%, further indicating that the explicit attention mechanism alone is insufficient to model the fine-grained texture features in expressions. The local statistical modulation and context texture enhancement mechanisms we introduced play an important role in improving the discriminative ability of the model.}

\par{\textbf{(2) Results on KDEF.} On the KDEF dataset, the TKFNet model we proposed also demonstrated leading performance, achieving a recognition accuracy rate of 92.04\%, as shown in Table 2, significantly surpassing the previous optimal methods, Dep-FER (91.20\%) and APViT (91.09\%). This result verifies the high robustness and good generalization ability of our method under standard experimental conditions. Although methods such as OCA-MTL and Latter-ofer have also achieved relatively high accuracy rates (89.04\% and 88.30\% respectively), there are still deficiencies in fine texture modeling and feature fusion. In contrast, TKFNet integrates local and global texture information more effectively and can distinguish the subtle differences in expressions more accurately, thus standing out among multiple benchmark models.}

\begin{table}[h]
	\centering
	\caption{Comparison with the state-of-the-art results on the RAF-DB dataset. The best results are in \textbf{BOLD}, and the second-best results are \underline{underlined}.}
	\label{tab:model_comparison}
	\footnotesize
	\begin{tabular}{@{} c c c @{}}
		\toprule
		\textbf{Model} & \textbf{Proc.} &\textbf{Acc. (\%)} \\ 
		\midrule
		HealthFERS\citep{c.bisogniImpactDeepLearning2022}&TII 22&\underline{82.63}\\
		Attention \citep{h.a.shehuAttentionBasedMethodsEmotion2024}&TETCI 24&81.09\\
		RGKT \citep{y.lvRelationshipGuidedKnowledgeTransfer2024}&TIP 24&72.34\\
		SqueezExpNet\citep{shahidSqueezExpNetDualstageConvolutional2023}&KBS 23&80.65\\
		DLP-CNN\citep{s.liReliableCrowdsourcingDeep2019}&TIP 19 &79.95\\
		MSAU-Net\citep{l.liangFineGrainedFacialExpression2021a}& TIFS 21&75.80\\
		EQCNN\citep{s.hossainDeepQuantumConvolutional2024}&TNSRE 24 &81.95\\
		TKFNet (OURS) & 2025 & \textbf{84.32} \\ 
		\bottomrule
	\end{tabular}
	
\end{table}

\begin{table}[!h]
	\centering
	\caption{Comparison with the state-of-the-art results on the KDEF dataset. The best results are in \textbf{BOLD}, and the second-best results are \underline{underlined}.}
	\label{tab:model_comparison}
	\footnotesize
	\begin{tabular}{@{} c c c @{}}
		\toprule
		\textbf{Model} & \textbf{Proc.}  & \textbf{Acc. (\%)} \\ 
		\midrule
		
		RUL \citep{zhangRelativeUncertaintyLearning2021} & NIPS 21  & 83.10 \\ 
		DML-Net \citep{liuDynamicMultichannelMetric2021}&INS 21 & 88.20 \\ 
		ECA \citep{zhangLearnAllErasing2022}& ECCV 22 & 88.00 \\ 		 
		OCA-MTL \citep{chenOrthogonalChannelAttentionbased2022}& PR 22& 89.04\\
		HealthFERS \citep{c.bisogniImpactDeepLearning2022}&TII 22 & 82.63\\
		SSA-Net \citep{liuJointSpatialScale2023}&PR 22&88.50\\
		Attention \citep{h.a.shehuAttentionBasedMethodsEmotion2024}& TETCI 24 &75.57\\
		EQCNN \citep{s.hossainDeepQuantumConvolutional2024}&TNSRE 24&81.95\\
		APViT \citep{f.xueVisionTransformerAttentive2023}&TAFFC 22&91.09\\			
		Latent-OFER \citep{i.leeLatentOFERDetectMask01} & ICCV 23 & 88.30 \\ 
		Dep-FER \citep{j.yeDepFERFacialExpression2024}& TAFFC 24  &\underline{91.20} \\ 

		TKFNet (OURS) & 2025 &\textbf{92.04} \\ 

		\bottomrule
	\end{tabular}
	\vspace{-3ex}
\end{table}

\subsection{Visualization}	
\par{Confusion matrix analysis is a key tool for evaluating the performance of a classification model. It provides a detailed breakdown of how well the model is able to distinguish between different classes by showing the true positives, false positives, true negatives, and false negatives for each emotion category. By examining the confusion matrix, we can identify specific classes where the model performs well and others where it may be making errors or misclassifications. In our experiments, we conduct confusion matrix analysis on the RAF-DB and KDEF datasets. It can be observed that on the RAF-DB dataset, our model performs well on expressions such as happiness and surprise, but shows poorer performance on fear and disgust. This is mainly due to the issue of class imbalance in the dataset. In contrast, the KDEF dataset is more balanced, resulting in better performance across various expressions, especially achieving up to 99\% accuracy on happiness and neutral.}
\begin{figure}[h] 
	\centering 
	\includegraphics[scale=0.48]{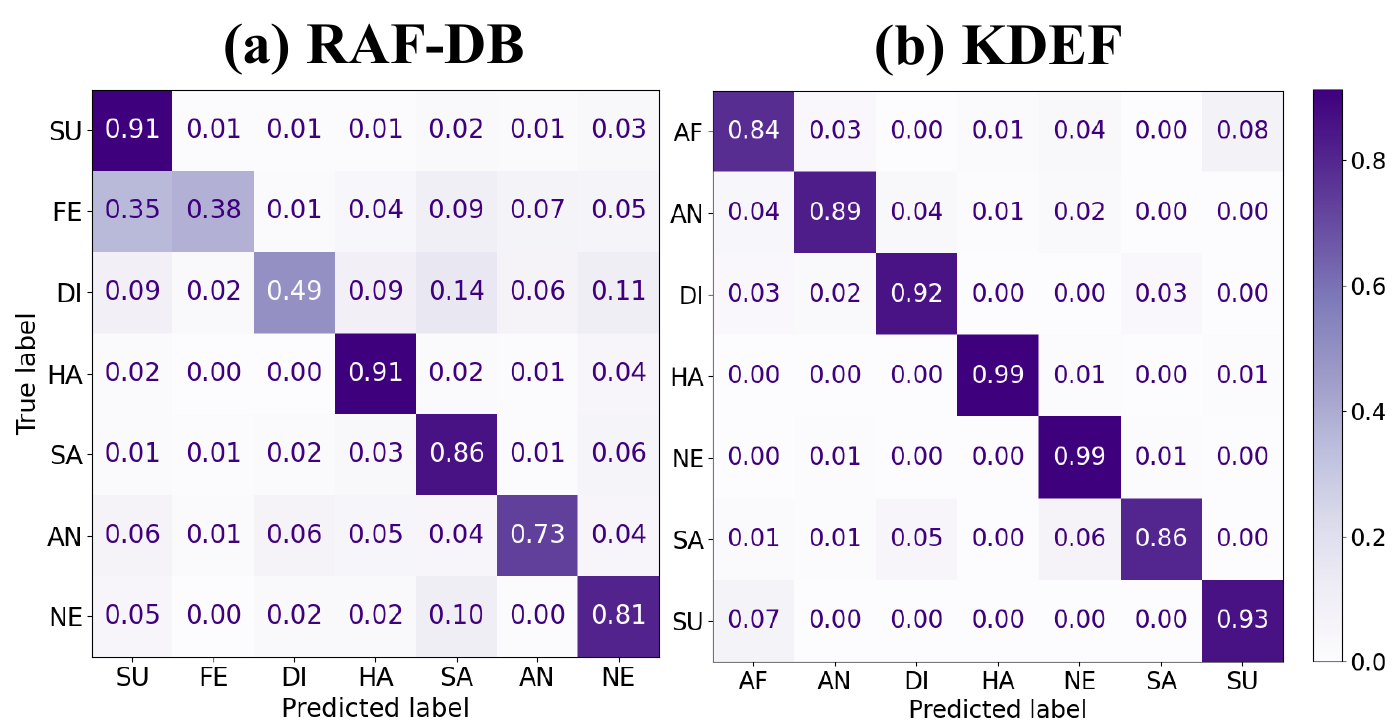}
	\caption{Confusion matrix of RAF-DB and KDEF.} 
	\label{FIG:5} 
\end{figure}
\section{Conclusion}
\par{In this paper, we proposed a novel facial expression recognition framework driven by Texture Key Driver Factors, which are essential in capturing subtle and discriminative facial texture variations. By introducing the TAFE, our model effectively enhances sensitivity to low-level semantic cues through multi-branch attention fusion. Moreover, the DCIF module adaptively refines the representation by selectively focusing on contextually relevant features. Experiments on RAF-DB and KDEF datasets demonstrate the robustness and superiority of our approach over state-of-the-art methods. The results validate the effectiveness of leveraging fine-grained texture patterns and contextual filtering in boosting expression recognition accuracy, particularly under challenging intra-class variation and inter-class ambiguity.}

\section*{Acknowledgments}
\par{Thanks for the support provided by MindSpore Community. All experiments proposed in this paper are implemented based on the mindspore framework.}

\bibliographystyle{unsrt}  

\bibliography{TKFNet}  

\end{document}